\documentclass{article}

\usepackage[utf8]{inputenc} 
\usepackage[T1]{fontenc}    
\usepackage[hyperfootnotes=false]{hyperref}       
\usepackage{url}            
\usepackage{booktabs}       
\usepackage{amsfonts}       
\usepackage{nicefrac}       
\usepackage{microtype}      
\usepackage{graphicx,graphics}
\usepackage{tikz}
\usepackage{svg}
\usepackage{amsmath,amssymb}
\usepackage{spconf,enumitem}
\usepackage{multirow}
\usepackage{xcolor}
\usepackage{mathcomEd}
\usepackage{siunitx}
\usepackage{booktabs,colortbl}
\usepackage{stfloats}
\usepackage{needspace}

\usepackage{array}
\usepackage{collcell}

\newcolumntype{H}{>{\collectcell\ignorecell}c<{\endcollectcell}@{}}
\newcommand{\ignorecell}[1]{}

\title{
The Path Matters: Evaluating Small Language Models \\ 
Beyond Answer Accuracy in KGQA
}

\name{Eduin E. Hernandez$^{*}$, Sergio A. Diaz, Luis F. Garcia, Nurassyl Askar, and Stefano Rini
\address{
NYCU, Taiwan
}
\thanks{
E.E.H., S.A.D., and S.R. are with National Yang Ming Chiao Tung University (NYCU), Taiwan.
L.F.G. and N.A. are independent researchers and performed part of this research while previously with NYCU.
$^{*}$Corresponding author: \texttt{eduin.ee08@nycu.edu.tw}.
\\
\indent This work has been submitted to the IEEE for possible publication. 
Copyright may be transferred without notice, after which this version may no longer be accessible.
}}

\begin{document}

\maketitle

\begin{abstract}
Small language models (SLMs) are increasingly paired with knowledge graphs (KGs), yet end-to-end KG question answering conflates graph access, search, navigation, reasoning, and answer generation.
This coupling makes it difficult both to determine whether an SLM can faithfully execute the reasoning path implied by a question and to attribute failures to navigation rather than to other stages of the pipeline.
We isolate this capability by employing the THESEUS navigation and traceability framework and using frozen, off-the-shelf SLMs as local action policies.
At each hop, the environment exposes the legal outgoing graph actions, and the model selects one executable graph action and decides whether to stop, without task-specific parameter updates, model-controlled beam search, or free-form answer generation.
This controlled setting allows us to evaluate terminal-answer accuracy with Hits@1 together with path fidelity, using Path Edit Distance (PED) as the primary trajectory metric.
Across the \textsc{Kinship} and \textsc{MQuAKE-ST} KGQAs, similarly sized local models differ substantially in answer accuracy and path fidelity, with the two metrics sometimes favoring different models.
This model-dependent behavior also extends to prompting, as a single demonstrated trajectory can improve or degrade navigation depending on the model.
These results motivate evaluating SLM graph reasoning beyond endpoint accuracy alone.
\end{abstract}


\section{Introduction}
\label{sec:introduction}
Knowledge graphs (KGs) provide explicit relational structure for multi-hop question answering, and recent language-model systems exploit this structure through graph retrieval, path search, planning, and grounded answer generation~\cite{sun2024thinkongraph,luo2024reasoningongraphs,sui2025fidelis}.
These systems demonstrate the value of graph scaffolding, but they make it difficult to determine whether a language model can choose the correct next graph action when search and answer synthesis are factored out.
This distinction is especially important for small language models (SLMs), where graph exploration has been identified as a major bottleneck inside a Think-on-Graph-style pipeline~\cite{cheng2025exploration}.
Meanwhile, recent work on compact KG agents has focused on improving performance through task-specific training rather than evaluating navigation ability of frozen models~\cite{jiang2025kgagent,xu2026graphwalker,sun2026sogr1}.

We study this question within the THESEUS formulation of multi-hop KGQA as question-conditioned graph navigation~\cite{hernandez2026theseus}, where the executed KG path is evaluated together with the terminal answer.
Our focus is a deliberately restricted setting: \emph{Can an off-the-shelf SLM faithfully execute a multi-hop reasoning path when the environment exposes the legal local graph actions and the model is responsible only for selecting the next executable action and deciding whether to stop?}
Isolating this capability reveals whether navigation failures arise from the model's local question-conditioned decisions rather than from retrieval, search, or answer synthesis, and provides a controlled baseline for measuring the value of additional reasoning mechanisms.
The controller fixes graph access, executes only legal edges, and uses the terminal entity as the answer, removing model-controlled candidate retrieval, pruning, beam/tree search, and free-form answer generation from the decision loop.
Related work has studied search-assisted KG reasoning and broader sequential graph navigation~\cite{sun2026search,ghandi2026graphwalk,margiotta2025wikigame}, whereas our setting isolates local executable graph-action selection and termination within path-annotated multi-hop KGQA.

Answer accuracy alone does not establish path fidelity: a model may reach a valid answer through a trajectory that differs substantially from the reference reasoning path~\cite{luo2024reasoningongraphs,li2025dog,sui2025fidelis}.
%
%
We therefore evaluate both ordered trajectory agreement, using Path Edit Distance (PED) over graph edges and order-invariant structural overlap, using $F1_{\mathrm{SG}}$ over traversed edges~\cite{hernandez2026theseus}.
PED serves as our primary trajectory metric.
We compare locally deployable SLMs under matched zero-shot and one-shot conditions on \textsc{Kinship} and \textsc{MQuAKE-ST}, finding that endpoint correctness and path fidelity can favor different models and that the effect of a single demonstrated trajectory is strongly model-dependent.

\textbf{Contribution.}
We provide a controlled empirical characterization of the local multi-hop graph-navigation ability of frozen, locally deployable SLMs under the THESEUS navigation and traceability framework, separating terminal-answer correctness from executed-path fidelity and examining the effect of a single inference-time demonstration.
%
%
As a secondary analysis, we use this controlled baseline in a compact ToG-style comparison on \textsc{MQuAKE-ST} to examine how explicit search and answer-generation scaffolding affect terminal accuracy and trajectory fidelity.

\section{Task Formulation and Evaluation}
\label{sec:task-formulation}

We build on the THESEUS formulation of multi-hop KGQA as question-conditioned graph navigation~\cite{hernandez2026theseus}.
Let a knowledge graph be $\mathcal{G}=(\mathcal{E},\mathcal{R},\mathcal{T})$, where $\mathcal{E}$ is the entity set, $\mathcal{R}$ is the relation set, and $\mathcal{T}\subseteq\mathcal{E}\times\mathcal{R}\times\mathcal{E}$ is the set of directed triplets.
Given a natural-language question $q$, a known topic entity $e_s$, and a valid answer set $\mathcal{A}(q)\subseteq\mathcal{E}$, the task is to execute a path
\begin{equation}
P=(e_0=e_s,r_1,e_1,\ldots,r_k,e_k)
\end{equation}
such that every $(e_{i-1},r_i,e_i)\in\mathcal{T}$ and the terminal entity $e_k\in\mathcal{A}(q)$.
As in THESEUS, the underlying reasoning length is not supplied to the agent.
Instead, questions are evaluated under a fixed maximum reasoning horizon $N$.
At step $t$, the legal outgoing graph-action set is
\begin{equation}
\mathcal{N}(e_t)=\{(r,e'):(e_t,r,e')\in\mathcal{T}\}.
\end{equation}
For our SLM instantiation, the presented graph actions are assigned local identifiers, and the model outputs a pair $(i_t,z_t)$ comprising a graph-action ID $i_t$ (or null) and a stop flag $z_t$.
If $i_t$ identifies a legal graph action, the controller deterministically executes the corresponding KG transition and appends the traversed triplet to the predicted path $\hat P(q)$.
If $z_t$ indicates stopping, the episode terminates after the selected transition.
When $i_t$ is null, the episode instead terminates at the current entity, and the null choice is valid only when stopping.
Otherwise, navigation continues until the maximum reasoning horizon $N$ is reached.
The terminal entity is returned as the predicted answer.

%
Our SLM policy produces a single trajectory rather than a ranked rollout set.
We therefore measure terminal-answer correctness with Hits@1, where the terminal entity must belong to $\mathcal{A}(q)$.
Because no ranked candidate set is produced, MRR is not defined in our evaluation.
For multi answer, reaching any valid answer counts as success.
%
%
For path fidelity, we follow the THESEUS evaluation protocol~\cite{hernandez2026theseus}.
PED is the Levenshtein distance between the ordered predicted and reference edge sequences, while $F1_{\mathrm{SG}}$ measures their order-invariant edge overlap.
Together, these metrics capture complementary aspects of trajectory fidelity: ordered path agreement and order-invariant edge overlap.
PED is our primary trajectory metric, with lower values indicating closer agreement and PED $=0$ corresponding to an exact reference-path match.
Together with Hits@1, these metrics provide a controlled measure of intrinsic local navigation against which additional search or task-specific adaptation can be evaluated.

\section{Experimental Setting}
\label{sec:experimental-setting}

\begin{table}[t]
\centering
\caption{
Datasets and evaluation settings. $\bar{k}$ denotes the mean reference path length over the evaluation questions.
}
\label{tab:eval_stats}
\fontsize{9pt}{11pt}
\setlength{\tabcolsep}{2pt}
\begin{tabular}{@{}lrrrrrr@{}}
\toprule
Dataset & $|\mathcal E|$ & $|\mathcal R|$ & $|\mathcal{T}|$ & Eval. Q & $\bar{k}$ & $N$ \\
\midrule
\textsc{Kinship}          & 24     & 12  & 112 & 101   & \num{2.623762376} & 3 \\
\textsc{MQuAKE-ST SA} & 38,516 & 665 & 724,141 & 1,504 & \num{2.513297872} & 4 \\
\textsc{MQuAKE-ST MA}  & 38,516 & 665 & 724,141 & 870   & \num{2.685057471} & 4 \\
\bottomrule
\end{tabular}
\end{table}

%
For the local-navigation experiments, all SLMs are evaluated under
the same navigation interface defined in Sec.~\ref{sec:task-formulation}.
At each step, the model observes the question $q$, topic entity $e_s$, current entity $e_t$, executed path history, and the presented legal outgoing graph actions.
We use tuple graph-action selection over $(e_t,r,e')$, with all models receiving the same indexed triplet representation of presented actions and the same structured response schema---\texttt{\{action: ID/null, stop: flag\}}.
Models are evaluated off the shelf with frozen parameters.
Optional model-specific thinking or reasoning modes are disabled to avoid introducing additional inference-time token and compute budgets as a confound.
Each navigation decision uses a single model generation, with no parse-retry or repair calls.

Our primary one-shot condition provides one complete solved trajectory from the training split as an in-context demonstration, showing the navigation state, available actions, and gold decision at each hop.
The zero-shot condition removes this demonstration while leaving the remaining navigation interface unchanged.
We use deterministic decoding with temperature zero, seed $42$, and a maximum response length of 64 tokens.
Code and evaluation scripts are publicly available.\footnote{\url{https://github.com/HalcyonSolutions/LLM_KGQA}}

We evaluate on the navigation-ready \textsc{Kinship} and \textsc{MQuAKE-ST} resources from THESEUS~\cite{hernandez2026theseus}, with \textsc{MQuAKE-ST} derived from the original \textsc{MQuAKE-CF} resource~\cite{zhong2023mquake}.
Each benchmark provides a fixed KG, explicit topic and valid answer entities, and annotated reasoning paths required for path-level evaluation.
Table~\ref{tab:eval_stats} summarizes the graph characteristics and evaluation subsets used in the present experiments.
We use action-display caps of 100 for \textsc{Kinship} and 200 for \textsc{MQuAKE-ST}.
For \textsc{MQuAKE-ST}, we evaluate two settings: Single Answer (SA) and Multi Answer (MA).

The evaluated model set comprises Gemma 4 E4B~\cite{gemmateam2026gemma4}, Granite 3.3 8B~\cite{ibm2025granite33}, Llama 3.1 8B~\cite{grattafiori2024llama3}, Ministral 3 8B~\cite{liu2026ministral3}, OLMo-3 7B~\cite{olmo2025olmo3}, Phi-4 Mini 3.8B~\cite{abouelenin2025phi4mini}, Qwen2.5 7B~\cite{yang2024qwen25}, and Qwen3 8B~\cite{yang2025qwen3}.
Llama 3.1, Ministral 3, Qwen2.5, and OLMo-3 use instruction-tuned local artifacts.
Llama 3.1, Ministral 3, and Qwen2.5 use Q4 artifacts.
Structured-output compatibility is an inclusion requirement of the controlled interface rather than an ablation dimension.\footnote{DeepSeek-R1 was excluded because it did not reliably satisfy the required structured response interface in our runs.}

\textbf{ToG-style search.}
To quantify the effect of explicit search, we additionally evaluate the three strongest models on \textsc{MQuAKE-ST SA}  using a ToG-style scaffold~\cite{sun2024thinkongraph}.
We preserve the same directed KG and maximum horizon $N=4$, restricting traversal to legal outgoing edges. 
The scaffold adds LLM-guided relation and entity pruning, global retention of $w$ candidate paths, accumulated-evidence sufficiency checking, and free-form answer generation.
These runs use the original ToG prompt templates rather than the structured zero- and one-shot navigation prompts used above.
We evaluate $w\in\{1,3\}$: $w=1$ provides a single-retained-path condition closest to our local-navigation setting, while $w=3$ follows the default ToG search width~\cite{sun2024thinkongraph}.
We report terminal-answer Hits@1, generated-answer accuracy, PED of the highest-ranked retained path, and mean LLM calls per question (Calls/Q).

\section{Results}
\label{sec:results}

\begin{table}[!t]
\centering
\caption{
One-shot navigation results on \textsc{Kinship} and \textsc{MQuAKE-ST} with structural calibration references reproduced from THESEUS~\cite{hernandez2026theseus}.
PED is the primary path-fidelity metric, and Calls/Q is the mean number of LLM invocations per question.
$\dagger$ denotes terminal answer Hits@1 undefined for the calibration references, and -- denotes not applicable.
}
\label{tab:main_results}
\fontsize{9pt}{11pt}
\setlength{\tabcolsep}{3pt}
\begin{tabular}{@{}lccHHcc@{}}
\toprule
Model / Reference & Hits@1 $\uparrow$ & PED $\downarrow$ & RED $\downarrow$ & $F1_{\mathrm{REL}}$ $\uparrow$ & $F1_{\mathrm{SG}}$ $\uparrow$ & Calls/Q \\
\midrule
\multicolumn{5}{l}{\textbf{(a) \textsc{Kinship}}} \\
\toprule
\emph{RW-AnsMC}             & $\dagger$ & 2.546 & 2.401 & 0.259 & 0.128 & -- \\
\emph{Shortest Path Oracle} & $\dagger$ & 2.099 & 1.871 & 0.350 & 0.220 & -- \\

\midrule

Gemma 4 E4B          & 0.911 & \textbf{0.158} & \textbf{0.129} & \textbf{0.979} & \textbf{0.958} & 2.663 \\
Granite 3.3 8B       & 0.554 & 0.842 & 0.752 & 0.848 & 0.768 & 2.861 \\
Llama 3.1 8B         & 0.436 & 0.832 & 0.743 & 0.860 & 0.783 & 3.000 \\
Ministral 3 8B       & \textbf{0.941} & 0.208 & 0.188 & 0.952 & 0.936 & 2.683 \\
OLMo-3 7B            & 0.337 & 1.307 & 1.149 & 0.790 & 0.638 & 2.931 \\
Phi-4 Mini 3.8B      & 0.347 & 1.446 & 1.188 & 0.705 & 0.530 & 2.495 \\
Qwen2.5 7B           & 0.683 & 0.614 & 0.495 & 0.901 & 0.825 & 2.743 \\
Qwen3 8B             & 0.822 & 0.446 & 0.347 & 0.921 & 0.860 & 2.594 \\
\midrule
\multicolumn{5}{l}{\textbf{(b) \textsc{MQuAKE-ST} Single Answer}} \\
\toprule
\emph{RW-AnsMC}             & $\dagger$ & 3.381 & 3.344 & 0.056 & 0.034 & -- \\
\emph{Shortest Path Oracle} & $\dagger$ & 1.670 & 1.418 & 0.503 & 0.380 & -- \\

\midrule

Gemma 4 E4B          & 0.728 & 0.680 & 0.583 & 0.847 & 0.790 & 2.604 \\
Granite 3.3 8B       & 0.373 & 2.001 & 1.811 & 0.590 & 0.489 & 3.337 \\
Llama 3.1 8B         & 0.351 & 2.227 & 2.110 & 0.634 & 0.561 & 3.965 \\
Ministral 3 8B       & \textbf{0.894} & \textbf{0.362} & \textbf{0.313} & \textbf{0.902} & \textbf{0.883} & 2.545 \\
OLMo-3 7B            & 0.173 & 2.322 & 2.197 & 0.480 & 0.410 & 3.349 \\
Phi-4 Mini 3.8B      & 0.297 & 1.521 & 1.420 & 0.560 & 0.513 & 1.751 \\
Qwen2.5 7B           & 0.666 & 0.955 & 0.840 & 0.768 & 0.716 & 2.849 \\
Qwen3 8B             & 0.824 & 0.520 & 0.453 & 0.869 & 0.843 & 2.438 \\
\midrule
\multicolumn{5}{l}{\textbf{(c) \textsc{MQuAKE-ST} Multi Answer}} \\
\toprule
\emph{RW-AnsMC}             & $\dagger$ & 3.406 & 3.370 & 0.066 & 0.044 & -- \\
\emph{Shortest Path Oracle} & $\dagger$ & 1.843 & 1.702 & 0.437 & 0.362 & -- \\

\midrule

Gemma 4 E4B          & 0.775 & 0.723 & 0.601 & 0.841 & 0.792 & 2.694 \\
Granite 3.3 8B       & 0.247 & 2.231 & 1.999 & 0.554 & 0.450 & 3.509 \\
Llama 3.1 8B         & 0.238 & 2.266 & 2.110 & 0.613 & 0.549 & 3.979 \\
Ministral 3 8B       & \textbf{0.849} & \textbf{0.601} & \textbf{0.491} & \textbf{0.861} & \textbf{0.812} & 2.657 \\
OLMo-3 7B            & 0.118 & 2.540 & 2.407 & 0.462 & 0.389 & 3.599 \\
Phi-4 Mini 3.8B      & 0.391 & 1.528 & 1.409 & 0.596 & 0.545 & 2.128 \\
Qwen2.5 7B           & 0.523 & 1.326 & 1.174 & 0.680 & 0.615 & 2.928 \\
Qwen3 8B             & 0.731 & 0.751 & 0.645 & 0.827 & 0.786 & 2.628 \\
\bottomrule
\end{tabular}
\end{table}

\textbf{Structural Calibration.}
Table~\ref{tab:main_results} reports one-shot navigation performance together with the structural calibration references.
The unbiased random walk represents unguided traversal, while the shortest-path oracle is given the valid answer set but not the annotated reasoning path and selects a shortest route to an answer.
The oracle is therefore not a bound on path fidelity, since the shortest route to a valid answer need not coincide with the reasoning path implied by the question.
Models with lower edit distances and higher overlap scores than the oracle follow the annotated reasoning structure more closely than can be achieved from answer knowledge and shortest-path efficiency alone.
Performance approaching the random-walk reference instead indicates weaker question-conditioned navigation.

\textbf{One-Shot Navigation Performance.}
Across the evaluated SLMs, terminal-answer accuracy and path fidelity vary substantially despite broadly similar deployment scales.
On \textsc{Kinship}, Ministral 3 achieves the highest Hits@1, whereas Gemma 4 achieves the strongest path fidelity across the reported metrics.
This shows that endpoint accuracy and path fidelity can favor different models.
On both \textsc{MQuAKE-ST} settings, however, Ministral 3 achieves the strongest performance on both answer accuracy and path fidelity.
Qwen3 and Gemma 4 form the next strongest group.
Several of the stronger SLMs also outperform the shortest-path oracle on the path-fidelity metrics, indicating closer agreement with the annotated reference reasoning structure than an answer-informed shortest route through the graph.
By contrast, weaker models move toward the structural calibration references, indicating less effective question-conditioned navigation.
Calls/Q should likewise be interpreted relative to the reference path length $\bar{k}$ rather than minimized independently, since low values can reflect premature stopping.

%
%

\begin{table}[!t]
\centering
\caption{
Zero-shot and one-shot navigation results on \textsc{Kinship} and \textsc{MQuAKE-ST}.
$\Delta$Calls/Q is the one-shot minus zero-shot mean number of LLM invocations per question.
Negative values indicate fewer calls under one-shot.
}
\label{tab:prompting_results}
\fontsize{9pt}{11pt}
\setlength{\tabcolsep}{4pt}
\begin{tabular}{@{}lccccc@{}}
\toprule
& \multicolumn{2}{c}{Hits@1 $\uparrow$}
& \multicolumn{2}{c}{PED $\downarrow$}
& \multirow{2}{*}{$\Delta$Calls/Q} \\
\cmidrule(lr){2-3}
\cmidrule(lr){4-5}
Model & 0-shot & 1-shot & 0-shot & 1-shot & \\
\midrule
\multicolumn{6}{l}{\textbf{(a) \textsc{Kinship}}} \\
\toprule
Gemma 4 E4B          & 0.653 & 0.911 & 0.634 & \textbf{0.158} & -0.337 \\
Granite 3.3 8B       & 0.634 & 0.554 & 0.861 & 0.842 & +0.119 \\
Llama 3.1 8B         & 0.455 & 0.436 & 0.713 & 0.832 & +0.000 \\
Ministral 3 8B       & \textbf{0.851} & \textbf{0.941} & \textbf{0.238} & 0.208 & -0.059 \\
OLMo-3 7B            & 0.436 & 0.337 & 0.950 & 1.307 & -0.059 \\
Phi-4 Mini 3.8B      & 0.396 & 0.347 & 1.416 & 1.446 & +0.129 \\
Qwen2.5 7B           & 0.485 & 0.683 & 0.911 & 0.614 & +0.287 \\
Qwen3 8B             & 0.703 & 0.822 & 0.653 & 0.446 & +0.119 \\
\midrule
\multicolumn{6}{l}{\textbf{(b) \textsc{MQuAKE-ST} Single Answer}} \\
\toprule
Gemma 4 E4B          & 0.588 & 0.728 & 1.111 & 0.680 & -0.578 \\
Granite 3.3 8B       & 0.417 & 0.373 & 1.623 & 2.001 & +0.725 \\
Llama 3.1 8B         & 0.257 & 0.351 & 2.170 & 2.227 & +0.002 \\
Ministral 3 8B       & \textbf{0.761} & \textbf{0.894} & 0.827 & \textbf{0.362} & +0.037 \\
OLMo-3 7B            & 0.239 & 0.173 & 2.211 & 2.322 & +0.023 \\
Phi-4 Mini 3.8B      & 0.124 & 0.297 & 1.658 & 1.521 & +0.370 \\
Qwen2.5 7B           & 0.590 & 0.666 & 1.205 & 0.955 & +0.265 \\
Qwen3 8B             & 0.681 & 0.824 & \textbf{0.769} & 0.520 & +0.162 \\
\midrule
\multicolumn{6}{l}{\textbf{(c) \textsc{MQuAKE-ST} Multi Answer}} \\
\toprule
Gemma 4 E4B          & 0.649 & 0.775 & 1.121 & 0.723 & -0.523 \\
Granite 3.3 8B       & 0.313 & 0.247 & 1.772 & 2.231 & +0.793 \\
Llama 3.1 8B         & 0.236 & 0.238 & 2.307 & 2.266 & +0.013 \\
Ministral 3 8B       & \textbf{0.708} & \textbf{0.849} & \textbf{0.907} & \textbf{0.601} & -0.037 \\
OLMo-3 7B            & 0.160 & 0.118 & 2.451 & 2.540 & +0.090 \\
Phi-4 Mini 3.8B      & 0.141 & 0.391 & 1.730 & 1.528 & +0.638 \\
Qwen2.5 7B           & 0.462 & 0.523 & 1.534 & 1.326 & +0.216 \\
Qwen3 8B             & 0.630 & 0.731 & 0.924 & 0.751 & +0.199 \\
\bottomrule
\end{tabular}
\end{table}

\textbf{Effect of One-Shot Demonstration.} 
Table~\ref{tab:prompting_results} compares zero-shot and one-shot navigation under the same evaluation interface.
The effect of a single demonstrated trajectory is strongly model dependent rather than uniformly beneficial.
Gemma~4, Ministral~3, and Qwen3 generally improve both terminal-answer accuracy and path fidelity under one-shot prompting, whereas Granite 3.3 degrades under the same intervention and OLMo-3 also weakens in several settings.
Other models exhibit mixed behavior.
In particular, Llama 3.1 can improve terminal-answer accuracy without a corresponding improvement in PED, showing that prompting can affect endpoint success and path agreement differently.
Changes in Calls/Q are similarly heterogeneous, with some models producing stronger trajectories using fewer calls and others requiring additional calls after the demonstration.
Overall, a single trajectory demonstration can affect terminal-answer accuracy, path fidelity, and stopping behavior differently across models.

\begin{table}[!t]
\centering
\caption{
ToG-style search on \textsc{MQuAKE-ST} Single Answer using the three
strongest local-navigation SLMs. 
%
Hits@1 evaluates the terminal entity and PED the trajectory of the highest-ranked retained path.
Gen.~Acc.\ evaluates the generated answer.
}
\label{tab:tog_search}
\fontsize{9pt}{11pt}
\setlength{\tabcolsep}{3pt}
\begin{tabular}{lccccc}
\toprule
\textbf{Model} &
$w$ &
\textbf{Hits@1} $\uparrow$ &
\textbf{Gen. Acc.} $\uparrow$ &
\textbf{PED} $\downarrow$ & 
\textbf{Calls/Q} \\
\midrule
Gemma 4 E4B   & 1 & \num[round-mode=places, round-precision=3, scientific-notation=fixed]{0.2892287234042553} & \num[round-mode=places, round-precision=3, scientific-notation=fixed]{0.4122340425531915} & \num[round-mode=places, round-precision=3, scientific-notation=fixed]{1.4414893617021276} & \num[round-mode=places, round-precision=3, scientific-notation=fixed]{5.380984042553192}\\
              & 3 & \num[round-mode=places, round-precision=3, scientific-notation=fixed]{0.15159574468085107} & \num[round-mode=places, round-precision=3, scientific-notation=fixed]{0.3071808510638298} & \num[round-mode=places, round-precision=3, scientific-notation=fixed]{1.9308510638297873}  & \num[round-mode=places, round-precision=3, scientific-notation=fixed]{5.96343085106383} \\
Ministral 3 8B & 1 & \num[round-mode=places, round-precision=3, scientific-notation=fixed]{0.3025265957446808} & \num[round-mode=places, round-precision=3, scientific-notation=fixed]{0.4833776595744681} & \num[round-mode=places, round-precision=3, scientific-notation=fixed]{1.4960106382978724} & \num[round-mode=places, round-precision=3, scientific-notation=fixed]{4.253324468085107} \\
              & 3 & \num[round-mode=places, round-precision=3, scientific-notation=fixed]{0.25} & \num[round-mode=places, round-precision=3, scientific-notation=fixed]{0.4900265957446808} & \num[round-mode=places, round-precision=3, scientific-notation=fixed]{1.8577127659574468} & \num[round-mode=places, round-precision=3, scientific-notation=fixed]{10.099734042553191} \\
Qwen3 8B      & 1 & \num[round-mode=places, round-precision=3, scientific-notation=fixed]{0.5465425531914894} & \num[round-mode=places, round-precision=3, scientific-notation=fixed]{0.7001329787234043} & \num[round-mode=places, round-precision=3, scientific-notation=fixed]{1.2892287234042554} & \num[round-mode=places, round-precision=3, scientific-notation=fixed]{6.294547872340425} \\
              & 3 & \num[round-mode=places, round-precision=3, scientific-notation=fixed]{0.3929521276595745} & \num[round-mode=places, round-precision=3, scientific-notation=fixed]{0.6934840425531915} & \num[round-mode=places, round-precision=3, scientific-notation=fixed]{1.6841755319148937}  & \num[round-mode=places, round-precision=3, scientific-notation=fixed]{12.987367021276595} \\
\bottomrule
\end{tabular}
\end{table}

\textbf{Effect of Explicit Search.}
Table~\ref{tab:tog_search} evaluates whether adding ToG-style search changes the behavior observed under local navigation.
Relative to Table~\ref{tab:main_results}, the $w=1$ setting yields lower Hits@1 and higher PED for all three models, despite requiring more LLM calls.
Notably, Ministral~3 and Gemma~4, which are among the strongest models in the isolated navigation setting, degrade substantially under the search scaffold, while Qwen3 remains comparatively stronger.
Increasing the width to $w=3$ further decreases Hits@1 and worsens PED across all models.
Generated-answer accuracy nevertheless exceeds terminal Hits@1 in every setting, showing that answer synthesis can recover correct answers from imperfect search trajectories.
Overall, under this matched directed setting, additional ToG-style search does not improve trajectory fidelity, and increasing search width provides no consistent benefit in generated-answer accuracy.
Because the scaffold jointly introduces prompting, relation and entity selection, path pruning, sufficiency checking, and answer generation, these results further motivate evaluating such components individually rather than attributing system-level behavior to search alone.

\section{Conclusion}
\label{sec:conclusion}

We evaluated frozen, off-the-shelf Small Language Models (SLM) as local graph-navigation policies under a controlled setting that fixes graph access, restricts execution to presented legal graph actions, and uses the terminal entity as the predicted answer.
Across \textsc{Kinship} and \textsc{MQuAKE-ST}, similarly sized models exhibit substantial differences in both terminal-answer accuracy and path fidelity.
Endpoint accuracy and path fidelity can rank models differently, as observed on \textsc{Kinship}, while the strongest model on \textsc{MQuAKE-ST} performs well on both criteria.
A single trajectory demonstration can also improve or degrade these measures depending on the model.
A complementary ToG-style evaluation shows that adding a multi-stage search-and-generation scaffold does not necessarily preserve the strong local-navigation behavior observed in isolation, while generated answers can partially mask trajectory failures.
These findings motivate component-wise evaluation of relation and entity selection, search control, stopping, and answer synthesis, as well as future study of task-specific adaptation.

\vfill\pagebreak
\clearpage

\section*{Acknowledgment}
This work is partially funded by the NSTC grant number 113-2923-E-A49-001 and by MARC, the MediaTek Advanced Research Center with grant number 114A540531.
OpenAI ChatGPT was used during manuscript preparation for language and clarity editing, wording refinement, and limited assistance with research ideation and code development.
All scientific decisions, experimental design, result interpretation, and final manuscript content were reviewed and verified by the authors, who take full responsibility for the work.

\section*{Compliance with Ethical Standards}
This study did not involve human or animal subjects, and no ethical approval was required.

{
\bibliographystyle{IEEEbib}
\bibliography{IEEEabrv, reference,related_work,models}
}

\end{document}